# A data-driven method for syndrome type identification and classification in traditional Chinese medicine

Nevin Lianwen Zhang, Ph.D.: The Hong Kong University of Science and Technology; E-mail:lzhang@cse.ust.hk

Chen Fu, M.D.: Dongfang Hospital, Beijing University of Traditional Chinese Medicine; E-mail: fuchen2003@163.com

Teng Fei Liu, Ph.D.: The Hong Kong University of Science and Technology; E-mail: liutf@cse.ust.hk

Bao Xin Chen, M.D.: Dongfang Hospital, Beijing University of Traditional Chinese Medicine; E-mail: chenbaoxin2008@163.com

Kin Man Poon, Ph.D.: The Hong Kong Institute of Education; E-mail: kmpoon@ied.edu.hk

Pei Xian Chen, B.S.: The Hong Kong University of Science and Technology; E-mail: pchenac@cse.ust.hk

Yun Ling Zhang, M.D.: Dongfang Hospital, Beijing University of Traditional Chinese Medicine; E-mail: yunlingzhang2004@163.com

Corresponding Author: Nevin L. Zhang, Department of Computer Science and Engineering, The Hong Kong University of Science and Technology; E-mail:lzhang@cse.ust.hk

**Abstract**

**Objective:** The efficacy of traditional Chinese medicine (TCM) treatments for Western medicine (WM) diseases relies heavily on the proper classification of patients into TCM syndrome types. We develop a data-driven method for solving the classification problem, where syndrome types are identified and quantified based on patterns detected in unlabeled symptom survey data.

**Method:** Latent class analysis (LCA) has been applied in WM research to solve a similar problem, i.e., to identify subtypes of a patient population in the absence of a gold standard. A widely known weakness of LCA is that it makes an unrealistically strong independence assumption. We relax the assumption by first detecting symptom co-occurrence patterns from survey data and use those patterns instead of the symptoms as features for LCA.

**Results:** The result of the investigation is a six-step method: Data collection, symptom co-occurrence pattern discovery, pattern interpretation, syndrome identification, syndrome type identification, and syndrome type classification. A software package called Lantern is developed to support the application of the method. The method is illustrated using a data set on Vascular Mild Cognitive Impairment (VMCI).

**Conclusions:** A data-driven method for TCM syndrome identification and classification is presented. The method can be used to answer the following questions about a Western medicine disease: What TCM syndrome types are there among the patients with the disease? What is the prevalence of each syndrome type? What are the statistical characteristics of each syndrome type in terms of occurrence of symptoms? How can we determine the syndrome type(s) of a patient?

1. **Introduction**

Traditional Chinese Medicine (TCM) is been increasingly used in healthcare in China and around the world as complementary or alternative to Western medicine (WM). A common practice is to divide the patients with a WM disease into several TCM syndrome types based on symptoms and signs (both referred as symptoms henceforth for simplicity), and to apply different TCM treatments to patients of different types. The efficacy of TCM treatments depends heavily on whether the classification is done properly.

The *TCM syndrome classification problem* associated with a WM disease consists of four subproblems: (1) What TCM syndrome types are there among the patients with the disease? (2) What is the prevalence of each syndrome type? (3) What are the characteristics of each syndrome type in terms of symptom occurrence probabilities? (4) How do we determine to the syndrome type(s) of a patient based on symptoms?

The syndrome classification problem is of fundamental importance to TCM research and clinic practice. As will be seen in Section 9, the problem has so far not been satisfactorily solved. In this paper we present a data-driven method for solving problem. The idea is to: (1) Conduct a cross-sectional survey of the patients with the WM disease and collect information about symptoms of interest to TCM; (2) Perform cluster analysis on the data and divide the patients into clusters based on symptom occurrence patterns; (3) Match the patient clusters with TCM syndrome types; (4) Use the statistical characteristics of the patient clusters to quantify the TCM syndrome types and to establish classification rules.

We will first explain the data analysis methods that this paper relies upon in Section 2. Then we will present our method for solving TCM syndrome classification in Sections 3 to 8. Related works will be discussed in Section 9, and conclusions drawn in Section 10. A data set on vascular mild cognitive impairment (VMCI) [1] will be used for illustration throughout the paper.

## 2. Technical Background

This paper builds upon two data analysis methods, namely latent class analysis (LCA) and latent tree analysis (LTA). They are based on probabilistic models that describe relationships among categorical variables. Some of the variables are observed, while the others are latent, that is, unobserved. In this section, we explain LCA and LTA in layman's term so that medical researchers without strong background in Statistics and Machine Learning can understand the key ideas.

### 2.1. Latent Tree Models and Latent Class Models

The models that we use are called *latent tree models* (LTMs). An LTM describes the relationship among a set of variables at two levels. At the qualitative level, it is an undirected tree where the observed variables are located at the leaf nodes, whereas the latent variables are located at the internal nodes. At the quantitative level, it describes the relationship between each pair of neighboring variables using a conditional probability distribution.

Figure 1(a) shows an example LTM taken from [2]. Qualitatively, it asserts that a student's Math grade (MG) and Science grade (SG) are influenced by his analytical skill (AS); his English grade (EG) and History grade (HG) are influenced by his literacy skill (LS); and the two skills are correlated. Here, the grades are observed variables, while the skills are latent variables.

For simplicity, assume all the variables have two possible values `low' and `high'. The dependence of MG on AS is quantitatively characterized by the conditional distribution P(MG|AS), which is also shown in Figure 1. It says that a student with high AS tends to get high MG and a student with low AS tends to get low MG. Similarly, the dependences of other grade variables on the skill variables are quantitatively characterized by the distributions P(SG|AS), P(EG|LS) and P(HG|LS) respectively. They are not shown to save space.

To specify the quantitative relationships among the latent variables, it is convenient to root the model at one of the latent variables and regard it as a directed model --- a tree-structured Bayesian network [3]. If we use AS as the root, then we need to provide, as given in Figure 1, the marginal distribution of the root P(AS) and the distribution P(LS|AS) of LS conditioned on its parent AS. If LS were chosen as the root instead, we would need to provide P(LS) and P(AS|LS). The choice of root does not matter because different choices give rise to equivalent directed models [4].

Latent tree models with a single latent variable are called *latent class models* (LCMs). Figure 1(b) shows an example LCM, where Intelligence is the sole latent variable. Qualitatively, it asserts that a student's grades in the four subjects are all influenced by his intelligence level.

Different models make different independence assumptions. In Figure 1(b), the four grade variables are assumed mutually independent conditioned on the latent variable Intelligence. This is known as the *local independence assumption*. Another way to state the assumption is that the correlations among the grade variables can be properly modeled using a single latent variable. In this sense, we also call it the *unidimensionality assumption*.

In Figure (a), the four grade variables are not assumed to be unidimensional. The correlations among the grade variables are modeled using two latent variables AS and LS. MG and SG are independent of each other conditioned on the latent variable AS, but EG and HG are not. Similarly, EG and HG are independent of each other conditioned on the latent variable LS, but MG and SG are not.

Historically, LCMs predate LTMs. LTMs are introduced as a generalization of LCMs in [4], where they are called hierarchical latent class models.

## 2.2. Latent Class Analysis

*Latent class analysis* (LCA) refers to the analysis of data using latent class models. As an example, consider a data set about the grades that students from a school obtain on the aforementioned four subjects. To perform LCA on the data, we assume there is a latent variable Y that is related to the grade variables as shown in Figure 2 (a). The task is to: (1) determine the number of possible values for Y, (2) determine the marginal distribution P(Y) and the distributions P(MG|Y), P(SG|Y), P(EG|Y) and P(HG|Y) of the grade variables conditioned on Y. The first task is known as *model selection*, while the second as *parameter estimation*.

In Statistics, the concept of *likelihood* measures how well a model fits data. In LCA,

probabilistic parameters are determined using the *maximum likelihood estimate (MLE) principle* [5]. The number of possible values for Y is often determined using the *Bayes Information Criterion (BIC)* [6]. The BIC score is likelihood plus a penalty term for model complexity. The use of BIC intuitively means that we want a model that fits data well, but do not want it to be overly complex.

Note that there are two equivalent versions of BIC in the literature that are negations of each other. In one version the BIC score takes positive values. Here we want to minimize the BIC score. In the other version it takes negative values. In this case, we want to maximize the BIC score.

In practice LCA is used as a tool for clustering discrete data [7]. Each value of the latent variable Y represents a probabilistic cluster of individuals and all the values collectively give a *partition* of all individuals. To determine the number of possible values for Y amounts to determine the number of clusters, and to determine the probabilistic parameters amounts to determine the statistical characteristics of the clusters.

## 2.3. Latent Tree Analysis

*Latent tree analysis* refers to the analysis of data using latent tree models. For a given data set, there are many possible LTMs. For example, one possible LTM for the imaginary student grade data is shown in Figure 2(b), another in Figure 2(a), and there are also other possible models. The task is to determine which model is the best for the data. Specifically, we need to determine: (1) the number of latent variables, (2) the number of possible values for each latent variable, (3) the connections among the latent and observed variables, and (4) the probability parameters. The model selection problem here is more difficult than in the case of LCA. It consists of the first three items.

Several algorithms have been developed for LTA [8]. Extensive empirical studies have been conducted where the different algorithms are compared in terms of the BIC scores of the models they obtain and running time [8,9]. The experiments indicate that the EAST (Extension-Adjustment-Simplification-until-Termination) algorithm [10] finds the best models on data sets with dozens to around one hundred observed variables, while the BI (Bridged-Islands) algorithm [9] finds the best models on data sets with hundreds to around one thousand observed variables. On data sets with dozens to around one hundred variables, BI is much faster than EAST, while the models it obtains are sometimes inferior. EAST is unable to deal with data sets with hundreds or more observed variables.

The LTM shown in Figure 2(b) can be viewed as two LCMs with their latent variables connected by an edge. In general, an LTM can be regarded as a collection of LCMs, where each LCM is based on a distinct subset of observed variables and the latent variables are connected to form a tree structure. As pointed out earlier, an LCM gives one probabilistic partition of data. Consequently, an LTM gives multiple partitions of data. Each partition is based primarily on a distinct subset of observed variables and the different partitions are correlated. For this reason, LTA is a tool for *multidimensional clustering* [10].

## 2.4. Software Tools

A software tool called Lantern has been developed to facilitate LCA and LTA. It is placed at [11]. The software is designed to run on desktop personal computers. The user can use it to analyze data using various algorithms, inspect the results, and perform further analyses to be described later in this paper. Separate implementations of the EAST and BI algorithms are also provided at [11] so that users can run data analysis on servers. On data sets that involve around 100 symptom variables and 1000 samples, EAST typically takes several days, while BI takes a few hours. We recommend that users run EAST on servers rather than on desktops.

## 3. Key Ideas of Method

LCA has been applied in WM research to identify subtypes of a patient population in the absence of a gold standard. The first applications appeared around 1990 and similar studies have increased sharply in recent years [12-15]. The studies primarily focus on infectious diseases, and also cover mental or behavior problems, diseases of the musculoskeletal system, disease of the digestive system, and neoplasms. Recently, LCA has also been used to identify TCM syndrome types among psoriatic patients [16].

When performing LCA, researchers begin by selecting, based on domain knowledge, the observed variables to be included in the analysis. Model parameters are estimated based on the maximum likelihood estimation principle and obtained using algorithms such as Expectation-Maximization and Newton-Raphson. The number of clusters is determined manually based on the BIC score and/or other goodness-of-fit measures. The output includes the prevalence (i.e., size) of each cluster, and a characterization of each clusters in terms of the distribution of the observed variables.

LCA makes the local independence assumption, which is often violated in practice. When this happens, the model obtained by LCA does not fit data well, and the estimates of subtype prevalence and other parameters are biased [12]. It can also lead to spurious clusters [17].

We propose to fit an LTM to data instead of an LCM. The fitting is automatically carried out by computers through a search procedure guided by the BIC score. In comparison with LCMs, LTMs are a much larger, and yet computationally manageable, class of models. Consequently, the model obtained by LTA would fit data much better than that obtained by LCA.

LTA yields a model with multiple latent variables. One example is shown in Figure 3. The structure of the model partitions all symptom variables into groups, with each group being all the variables directly connected to a latent variable. The variables in each group are mutually independent given the latent variable. So, LTA partitions all the symptom variables into unidimensional subsets.

Intuitively, each of the latent variable represents one latent aspect of data. As will be seen in Section 4, a latent aspect manifests as either probabilistic co-occurrence of a group symptoms, or probabilistic mutual exclusion of two subgroups of symptoms, with symptoms in each subgroup tend to co-occur.

Each of the latent variables also gives a partition of data. However, the patient clusters in such partitions usually do not correspond to TCM syndrome types. There are two reasons. First, the definition of a TCM syndrome type typically requires information from multiple latent aspects of data. Second, one latent aspect of data might be related to multiple syndrome types. Specifically, two symptoms that tend to be mutual exclusive are usually caused by different syndrome types. Two symptoms that tend to co-occur might be caused by a single syndrome type or by two different syndrome types that co-occur.

To identify the cluster of patient that corresponds to a TCM syndrome type, we first select a set of symptom variables based on domain knowledge and perform cluster analysis to group patients based on those variables. However, we do not always use the symptom variables themselves as features for the cluster analysis. If two or more symptom variables are from the same latent aspect of data, we "combine" them into one latent feature, and use the latent feature instead of the symptom variables themselves. The objective here is to relax the local independence assumption.

Figure 4 shows the model that we use to identify the cluster of patients that correspond to the symptom type Phlegm-Dampness. According to domain knowledge, the symptoms insomnia and dreamfulness are both related to Phlegm-Dampness, and hence are included in the model. According to the results of LTA (Figure 3), they are from the same latent aspect of data, and hence are not used as features directly. Instead, they are "combined" into one latent feature and the latent feature is connected to the clustering variable Z. Given Z, the two symptom variables are not mutually independent, and hence the local independence assumption is relaxed.

We call our method for solving the TCM syndrome classification problem the *LTA method* because of its reliance on latent tree analysis. The details of the method are presented in the next five sections.

## 4. Symptom Pattern Discovery

We illustrate the LTA method using a data set on vascular mild cognitive impairment (VMCI). The data set involves 803 patients and 93 symptoms. Detailed information about the data set and the data collection process is given in [1].

The first step of the method is to perform LTA on the VMCI data. This is done using the EAST algorithm. The structure of the resultant model is shown in Figure 3 and the BIC score of is -27,824. In contrast, the BIC score of the model obtained by LCA is -29,566. The model obtained by LTA fits the data much better than the model obtained by LCA.

In Figure 3, the variables labeled with English phrases are symptom variables, which are from the data set. The Y-variables are latent variables, which are introduced during data analysis. The integer next to a latent variable is the number of its possible values. For example, Y01 has 2 possible values, whereas Y15 has 3.

The widths of the edges indicate strengths of correlations between neighboring variables. We see that Y01 is strongly correlated with the symptom variables sallow complexion, asthenia of defecation, and dry stool or constipation, and weakly related to clear profuse urination. Similarly, Y08 is strongly correlated with the latent variables Y09 and Y12, and weakly related to Y04 and Y13.

The latent variables reveal symptom co-occurrence patterns and symptom mutual-exclusion patterns hidden in the data. Each of those patterns can be intuitively understood as the manifestation of a latent aspect of the data. In the rest of this section, we examine a few of the patterns.

Each of the latent variables represents a probabilistic partition of the patients included in the data set. For example, Y06 has two possible values and hence partitions the patients into two clusters, which are denoted as Y06=s0 and Y06=s1 respectively. Table 1 shows the information about the partition compiled using the Model Interpretation function of the Lantern software.

The symptoms in the table are those that are directly connected to Y06 in Figure 3. They are sorted in descending order of their mutual information with the partition, or, equivalently, with the variable Y06. *Mutual information (MI)* measures the amount of information that a symptom variable contains about the partition, or vice versa. It is closely related to the difference in occurrence probabilities of the variables in the two clusters. The larger the mutual information, the larger the difference in occurrence probabilities, and the more important the variable is for distinguishing between the two clusters. Among the two symptoms in Table 1, thick tongue fur is more important because its occurrence probabilities in the two clusters differ more than those of greasy tongue fur.

We see that the two clusters consist of 79% and 21% of the patients respectively.

The symptoms occur with higher probabilities in the cluster Y06=s1, and with lower probabilities in the cluster Y06=s0. Those indicate that the two symptoms are positively correlated. When one symptom occurs, the other also tends to occur. In other words, the two symptoms tend to co-occur. In this sense, Y06 reveals the probabilistic *co-occurrence* of the two symptoms. This is the *statistical meaning* of Y06.

Table 2 shows the information about that partition given by the latent variable Y20. It reveals the probabilistic co-occurrence of the three symptoms fat tongue, tongue with ecchymosis, and tooth-marked tongue.

Table 3 shows the information about the partition given by the latent variable Y12. In Figure 3, Y12 is directly connected to the two symptom variables slippery pulse and thin pulse. In the cluster Y12=s0, slippery pulse occurs with high probability and thin pulse does not occur at all. In the cluster Y12=s1, on the other hand, slippery pulse occurs with low probability and thin pulse occur with high probability. Those indicate that the two symptoms are negatively correlated. When one symptom occurs, the other tends not to occur. In this sense, Y12 reveals the *probabilistic mutual exclusion* of slippery pulse and thin pulse.

Table 4 shows the information about the partition given by yet another latent variable Y25. It reveals the probabilistic mutual exclusion of two groups of symptoms: (1) insomnia and dreamfulness, and (2) flushed face. Moreover, it indicates that two symptoms in the first group tend to co-occur.

In general, a latent variable reveals either that a group of symptoms tend to co-occur, or that two subgroups of symptoms tend to be mutual exclusive, with symptoms in each group tend to co-occur. An examination of all the latent variables in Figure has found that Y02, Y05, Y09, Y10, Y12, Y14, Y15 and Y25 fall into the second category, whereas the other latent variables fall into the first category. The details are given in [1].

## 5. Pattern Interpretation

Our objective is to, based on the VMCI data and domain knowledge, identify patient clusters that correspond to TCM syndrome types, and to quantify the syndrome types using the statistical characteristics of the patient clusters. We need to answer three questions to begin with: What are the target syndrome types? What information in the data should we use for each target syndrome type? Is the information sufficient?

One way to find the answers for the questions is to examine the symptoms in the data one by one and, for each symptom, list all the syndrome types that can, according to domain knowledge, lead to the occurrence of the symptom. This way, a list of syndrome types is associated with each symptom. All the syndrome types that appear in the lists are the target syndrome types. For each target syndrome type, all symptoms in the data set that it can cause to occur constitute the information we should use for the syndrome type. The information is sufficient if all key manifestations of the syndrome type are covered.

Examining the symptoms one by one is tedious. We examine them in groups instead. As seen in the previous section, LTA has discovered symptom co-occurrence patterns and thereby divided the symptoms into groups. We examine the patterns one by one. For each pattern, we identify the syndrome types that, according to domain knowledge, can lead to the co-occurrence of the symptoms in the pattern. This step is hence called *Pattern Interpretation*.

We begin with Y06, which reveals the probabilistic co-occurrence of thick tongue fur and greasy tongue fur. To determine the TCM connotation of the pattern, we ask this question: What TCM syndrome type or types can bring about the co-occurrence of the two symptoms? According to domain knowledge [e.g., 18], the answer is Phlegm-Dampness. Hence, Phlegm-Dampness is the interpretation for the pattern.

Y20 reveals the probabilistic co-occurrence of the three symptoms fat tongue, tongue with ecchymosis, and tooth-marked tongue. According to domain knowledge, the first and the third symptoms can be caused by either Phlegm-Dampness or Qi Deficiency, and the second symptom can be caused by Blood Stasis. Therefore, the co-occurrence of the three symptom can be due to either the co-occurrence of Phlegm-Dampness and Blood Stasis, or the co-occurrence of Qi Deficiency and Blood Stasis. We regard Phlegm-Dampness and Qi Deficiency as (two alternative) *primary interpretations* of the pattern, because they explain the leading symptom fat tongue. In contrast, Blood Stasis is regarded as a *secondary interpretation* of the pattern.

If a latent variable reveals the mutual exclusion of two groups of symptoms, the two groups are interpreted separately. For example, Y12 reveals the probabilistic mutual exclusion of the two symptoms slippery pulse and thin pulse. Slippery pulse can be explained by Phlegm-Dampness, while thin pulse can be explained by either Qi Deficiency or Blood Deficiency. Y25 reveals the probabilistic mutual exclusion of two groups of symptoms: (1) insomnia and dreamfulness, and (2) flushed face. The first group can be explained by any of several syndrome types, namely Yin Deficiency, Fire-Heat, Blood Deficiency, Qi Deficiency, and Phlegm-Dampness. The second group can be explained by either Fire-Heat or Yin Deficiency.

Note that the interpretation of a co-occurrence pattern is about identifying an appropriate explanation for the pattern. It is not about determining the syndrome type of a patient given that the pattern is present. To interpret the co-occurrence of insomnia and dreamfulness, for instance, the correct question to ask is: What syndrome types can lead to the occurrence of the two symptoms? There might be multiple possible answers for such a question as indicated above. The wrong question is: What is the syndrome type of a patient if he has the two symptoms? This question cannot be answered because it is not possible to determine the syndrome type of a patient based only on the two symptoms insomnia and dreamfulness.

Pattern interpretation requires domain knowledge and hence expert judgments. It is possible that different experts interpret the same pattern in different ways. If that happens, the issue can be resolved through discussions among TCM researchers, for example, by following the Delphi method [19]. Pattern interpretation is where TCM researchers need to spend the most efforts when using the LTA method.

## 6. Target Syndrome Type Identification

The pattern interpretation process gives rise to a list of syndrome types. Each of them is a potential target for syndrome quantification. The next step is to determine what symptoms in the data set should be used to quantify the syndrome types and whether the information is sufficient.

One of the potential target syndrome types identified in the previous section is Phlegm-Dampness. According to the discussions, the following symptoms should be used when quantifying the syndrome type: Y06 (thick tongue fur, greasy tongue fur), Y20 (fat tongue, tooth-marked tongue), Y12 (slippery pulse), and Y25 (insomnia, dreamfulness).

In [1], all latent variables in Figure 3 are examined and interpreted. The interpretation suggest that the following information should also be included when quantifying Phlegm-Dampness: Y11 (sticky or greasy feel in mouth), Y21 (dizziness, head feels as if swathed, distending headache, nausea or vomiting), Y03 (dizzy headache), Y14 (thirst desire no drinks), Y16 (urinary incontinence), and Y19 (expectoration).

There are totally sixteen symptoms. They are about various manifestations of Phlegm-Dampness. Together, they cover all major aspects of the impacts of Phlegm-Dampness. It is therefore concluded that Phlegm-Dampness is well supported by the data and it can be a target syndrome type for quantification.

## 7. Syndrome Quantification

To quantify Phlegm-Dampness, we perform cluster analysis on the patients based on the sixteen symptom variables and thereby identify a patient cluster that corresponds to the syndrome type. The size of the patient cluster is then regarded as a measurement of the prevalence of Phlegm-Dampness, and occurrence probabilities of the sixteen symptoms in the cluster are used as the definition of Phlegm-Dampness.

The cluster analysis is carried out using the model shown in Figure 4. The latent variable Z at the top represents the patient clusters to be identified. The features are from various latent aspects of the data. If there is only one symptom variable from an aspect, it is connected to Z directly. If there are multiple symptom variables from an aspect, they are connected to Z via an intermediate latent variable. As pointed out in Section 3, the use of the model in Figure 4 instead of the latent class model with all the symptom variables as feature relaxes the local independence assumption.

The analysis divides the patients into clusters by jointly considering information from various aspects of data. Hence it is called *joint clustering* and Z is called the *joint clustering variable*. The computational task is to determine the number of possible values for Z and the probability parameters. This is done using Lantern through a procedure similar to standard LCA.

The result of the joint clustering is a partition with three patient clusters, which are denoted as Z=s0, Z=s1 and Z=s2 respectively. Information about the partition is given in Table 5. There are sixteen symptom variables in the joint clustering model. Only seven of them are shown in Table 5 for simplicity. The others are pruned by Lantern based on cumulative information coverage. Specifically, Lantern sorts the sixteen variables in descending order of their mutual information with the joint clustering variable Z. The *cumulative information coverage (CIC)* of a variable in the ordered list is a ratio, where the numerator is the amount of information about the partition contained in the variable and in all those before it, and the denominator is the amount of information about the partition contained in all the variables [10]. It exceeds 95% at the variable dizziness. Hence, we conclude that the first seven variables are sufficient for characterizing the differences among the three clusters.

In Table 5, we see that the symptoms, especially the first three, occur with much higher probabilities in the clusters Z=s1 and Z=s2 than in the cluster Z=s0. We hence interpret Z=s1 and Z=s2 as two subtypes of Phlegm-Dampness and Z=s0 as the class of patients without Phlegm-Dampness. In this paper, our goal is to quantify the syndrome type Phlegm-Dampness and we are not interested in exploring it's subtypes. Hence, we merge the two clusters Z=s1 and Z=s2 into one cluster Z=s12, and interpret Z=s12 as the class of patients with Phlegm-Dampness.

If we accept the above interpretations and if the patients surveyed are a representative sample of the general VMCI population, then we have obtained a quantification of Phlegm-Dampness for VMCI. According to the quantification results (columns 2 and 5 of Table 5), 58% of the VMCI patients have Phlegm-Dampness and 42% of them do not. In the class of Phlegm-Dampness (Z=s12), greasy tongue fur occurs with probability 0.80, slippery pulse occurs with probability 0.60, and so on. In the class of non- Phlegm-Dampness (Z=s0), the symptoms occur with much lower probabilities.

## 8. Syndrome Classification

In the previous two sections, we have discussed how to identity and quantify TCM syndrome types based on domain knowledge and symptom co-occurrence patterns discovered in data. In this section, we discuss the last step of the LTA method, which is to derive classification rules for the syndrome types.

### 8.1 Model-Based Classification

As seen in the previous section, quantitative characterizations of TCM syndrome types are obtained using joint clustering models such as the one shown in Figure 4. For the following discussions, we consider a general joint clustering model. Denote the symptom variables the model as $X_1, ..., X_n$ and the joint clustering latent variable as $Z$. Each symptom variable has two possible states 0 and 1, which represent the absence and presence of the symptom respectively. The variable Z has two or more states. Let *s* be a state of *Z* that is interpreted as a syndrome type and use ~*s* to denote the other state(s). The problem is to determine whether *Z=s* or *Z=~s* based on the values of the symptom variables.

The problem is simple in theory. All we need to do is to compute the posterior distribution $P(Z|X_1,\ldots,X_n)$ of *Z* given the values of the symptom variables, and conclude that *Z=s* if its posterior probability is higher than that of *Z=~s*, that is,

$$P(Z = s|X_1,\ldots,X_n) > P(Z = \sim s|X_1,\ldots,X_n). \qquad (1)$$

This method is called *model-based classification*. Although conceptually simple, it lacks operability. It is unrealistic to expect physicians to do probability calculations in the clinic setting. It is of course possible to write a software tool for the physicians to use as a black box. However, it might be difficult for patients and physicians to trust black boxes.

### 8.2 Scored-Based Classification Rules

For the sake of operability, we derive a score-based classification rule to approximate model-based classification. Scores are associated with the symptoms. The total score for a patient is calculated based on the presence and absence of the symptoms. When the total score exceeds a threshold, the patient is classified into the class Z=s.

We start by rewriting inequality (1) into the following equivalent form using Bayes rule:

$$P(Z = s)P(X_1,\ldots,X_n|Z = s) > P(Z = \sim s)P(X_1,\ldots,X_n|Z = \sim s).$$

To obtain a score-based classification rule, we assume that the symptom variables are mutually independent given Z=s or Z=~s. Strictly speaking, the assumption is not true. In Figure 4, for instance, the two variables greasy tongue fur and thick tongue fur are not independent of each other given Z. Hence approximations are introduced here and their impacts will be assessed later. The assumption allows us to rewrite the inequality further as follows:

$$P(Z = s)P(X_1|Z = s)\ldots P(X_n|Z = s) > P(Z = \sim s)P(X_1|Z = \sim s)\ldots P(X_n|Z = \sim s).$$

By taking logarithm of both sides and re-arranging the terms, we get:

$$\log \frac{P(X_1|Z = s)}{P(X_1|Z = \sim s)} + \ldots + \log \frac{P(X_n|Z = s)}{P(X_n|Z = \sim s)} > \log \frac{P(Z = \sim s)}{P(Z = s)} \quad (2)$$

By subtracting a constant from both sides of inequality (2), we get:

$$\sum_{i=1}^{n} \log(\frac{P(X_i|Z = s)}{P(X_i|Z = \sim s)} / \frac{P(X_i = 0|Z = s)}{P(X_i = 0|Z = \sim s)}) > \log \frac{P(Z = \sim s)}{P(Z = s)} - \sum_{i=1}^{n} \log \frac{P(X_i = 0|Z = s)}{P(X_i = 0|Z = \sim s)} \quad (3)$$

Two technical remarks are in order. First, the choice of base for logarithm does not matter. We use base 2. Second, the probability values might be 0 sometimes. We deal with this issue by smoothing. The formula for calculating the conditional distribution $P(X_i|Z)$ is $P(X_i|Z) = \frac{P(X_i, Z)}{P(Z)}$. Smoothing means to change the formula to $P(X_i|Z) = \frac{P(X_i, Z) + c}{P(Z) + |X_i|c}$, where $|X_i|$ is the number of possible values of $X_i$ and $c$ is the smoothing parameter. The smoothing parameter is to be determined by the user. In Lantern, it is set at 0.000001 by default.

Note that on the left hand side of inequality (3), there is one term for each symptom variable. We regard it as the score for that symptom. To be more specific, the score for symptom variable $X_i$ is:

$$\log(\frac{P(X_i|Z=s)}{P(X_i|Z=\sim s)} / \frac{P(X_i=0|Z=s)}{P(X_i=0|Z=\sim s)})$$

Theoretically, there are two scores for each symptom variable, one for $X_i=0$ (absence of the symptom), and another for $X_i=1$ (presence of the symptom). However, the score for $X_i=0$ is always 0. So, there is in effect only the score for $X_i$, i.e., the score for the presence of the symptom.

The term on the right hand of (3) is regarded as the threshold. Under this interpretation, the inequality becomes a classification rule and it is an approximation to the rule given by inequality (1).

Table 6 shows the classification rule for Phlegm-Dampness derived from the joint clustering model in Figure 4. We see that the score for greasy tongue fur is 7.1, the score for slippery pulse is 2.1, and so on. The threshold is 4.2.

To use the classification rule on a patient, we need to examine and see whether the symptoms in the table are present. The patient gets one score for a symptom if the symptom is present. If the total score exceeds the threshold 4.2, the patient is classified into the Phlegm-Dampness class (*Z=s12*). Otherwise, the patient is classified into the non-Phlegm-Dampness class (*Z=s0*).

The classification rule is an approximation of model-based classification. How accurate is the approximation? To answer this question, we applied both methods on the patients in the VMCI data set. It turns out that the rule classifies 96.9% of the patients the same way as model-based classification. Therefore, the accuracy of the rule is 96.9%, as indicated at the bottom of the "accuracy" column of Table 6.

**8.3 Understanding the Scores**

It is important to realize that the scores in Table 6 are calculated from the probability values in Table 5, and they have an intuitive interpretation. To see this, note that:

$$\log(\frac{P(X_i=1|Z=s)}{P(X_i=1|Z=\sim s)} / \frac{P(X_i=0|Z=s)}{P(X_i=0|Z=\sim s)}) = \log(\frac{P(X_i=1|Z=s)}{P(X_i=0|Z=s)} / \frac{P(X_i=1|Z=\sim s)}{P(X_i=0|Z=\sim s)})$$

The first fraction on the right hand side is the odds for observing $X_i =1$ in the class $Z=s$, while the second term is the odds for observing $X_i =1$ in the class $Z=\sim s$. So, the score is simply a *log odds ratio*, a standard term in Statistics to quantify how strongly the presence or absence of one property (the symptom $X_i$) is associated with the presence or absence of another property (whether the patient is in the class $Z=s$). The score for a symptom is large if it occurs with high probability in $Z=s$ and low probability in $Z=\sim s$.

As an example, consider how the score for the symptom greasy tongue fur is computed from Table 5. According to Table 5, the symptom occurs in the Phlegm-Dampness class with probability *0.80* and does not occur with probability *0.20 (=1-0.80)*. So, the odds to observe symptom in the Phlegm-Dampness class is *0.80/0.20=4.0*. On the other hand, the symptom occurs in the non-Phlegm-Dampness with probability *0.03* and does not occur with probability *0.97*. So, the odds to observe the symptom in the non-Phlegm-Dampness class is *0.03/0.97≈0.03*. Because the odds to observe the symptom is higher in the Phlegm-Dampness class than in the non-Phlegm-Dampness, it is positive evidence for Phlegm-Dampness. Its score is the logarithm of the odds ratio, that is, $log_2(4.0/0.03)≈7.1$.

The threshold 4.2 is also computed from probability values in Table 5. Intuitively, it is the total strength of the evidence for non-Phlegm-Dampness when all the symptoms are absent.

Note that the position of a symptom in the classification rule is determined not only by its score, but also by how often it occurs. For example, the score for the symptom sticky or greasy feel in mouth is 2.8, which is higher than the score for the symptom slippery pulse (2.1). However, the former symptom occurs with lower probability in the data than the latter symptom and is hence applicable to a smaller fraction of the patients. Consequently, it is placed behind slippery pulse in the rule.

**8.4 Simplification of Classification Rules**

Several symptoms Table 6 have low scores. We can consider eliminating such symptoms so as to simplify the classification rule.

The elimination of symptoms from a classification rule would affect its accuracy. The impact needs to be assessed before the elimination actually takes place. The Lantern software has a function to facilitate this operation. To illustrate the function, we consider eliminating several symptoms at the bottom of Table 6. The accuracies of the simplified rules are shown in the "accuracy" column. The number 0.969 at the bottom is the accuracy for the rule with no symptom removed; the second last number, which happens to be also 0.969, is the accuracy for the rule with the last symptom removed; and so on. We see that, if we keep the top 8 symptoms and remove the bottom 8 from the rule, the rule becomes much simpler and its accuracy is still high (0.967). Consequently, we recommend the rule with the first 8 symptoms as the final rule. The threshold for the simplified rule is 3.7.

**8.5 Integer-Valued Classification Rules**

In the literature it is customary to present classification rules using integer symptom scores and threshold [20]. However, the classification rules produced by the LTA method is real-valued. It is possible to convert real-valued rules into integer-valued rules [21].

A real-valued rule can be converted into an integer-valued rule by simply applying rounding to the symptom scores and the threshold. Obviously, rounding would affect the accuracy of the rule. To minimize the impact, we can multiply all the scores and the threshold by a scaling factor and then applying rounding. This operation is supported by the Lantern software.

There are two drawbacks with integer-valued classification rules. First, the symptom scores and the threshold no long have the semantics as described in Section 8.3.   Second, different researchers might use different scaling factors even if they work on the same problem. This renders their results not comparable. For the field to move forward, it is important that results from different research groups are comparable. For this reason, we recommend not to round up real-valued rules. For the same reason, it is not advisable to enforce that all classification rules have the same threshold as suggested in [20].

## 9. Discussions

Previous research on the TCM syndrome classification problem has been conducted in three directions. The first line of research is known as *TCM Syndrome Essence Study (Zheng Shi Zhi Yan Jiu)*. The objective is to identify biomarkers that can be used as gold standards for TCM syndrome classification. Despite extensive efforts by many researchers over a long period of time, little success has been achieved. When reflecting on the efforts in this direction, Zhao wrote in 1999 that "Syndrome Essence Study has accumulated a lot of experimental data in the past 40 years. Against the original research objectives, all the data have brought us are confusion and perplexity." [22]

In the second line of research, panels of experts are set up to establish syndrome classification standards for various WM diseases [e.g., 23,24]. We call such efforts *Panel Standardization*. A standard set up this way includes a list of syndrome types that are deemed present among the patients with a WM disease. It also provides, for each syndrome type, a list the symptoms that are likely to occur, and highlights the key symptoms for classification. There are typically no quantitative information about syndrome prevalence and symptom occurrence probabilities, and no clearly specified classification rules. Moreover, the standards are based on subjective opinions of experts, and different panels might produce different standards [25].

The third line of research is based on labeled clinic data [e.g. 26,20,27]. In such a study, the patients of a WM disease are surveyed and information about symptom occurrence is collected. The patients are examined by TCM physicians and their syndrome types are determined. In other words, the data have syndrome class labels and hence the data are labeled. Statistics are then calculated based on the labels to determine syndrome prevalence and symptom occurrence probabilities for each syndrome type. Classification rules are established using statistical and machine learning techniques such as regression, neural networks and support vector machines [28]. The conclusions are summaries of the behaviors of the TCM physicians who participate in the studies. Hence we call such work *Supervised Quantification*.

Syndrome Essence Study would provide the strongest evidence for TCM syndrome classification, but it has not been successful. Panel Standardization and Supervised Quantification are based on weak evidence, and the results depend heavily on the experts who participate in the research work.

In this paper, we have presented a novel approach to TCM syndrome classification, namely the LTA method. The method starts with data on symptom occurrence. There are no judgments about syndrome types and hence the data are unlabeled. The data are analyzed using latent tree models to identify symptom co-occurrence patterns, and the patterns are used to identify patient clusters that correspond to syndrome types. The statistical characteristics of the patient clusters are then used to quantify the syndrome types and establish classification rules. This new approach can also be called *Unsupervised Quantification*. Although domain knowledge is required when interpreting the patterns discovered, the approach places TCM syndrome classification on the basis of stronger evidence than Panel Standardization and Supervised Quantification. The evidence might not be as strong as what one would get from Syndrome Essence Study, but the latter has not been successful so far. So, the LTA method represents the best of what can be done for the time being.

In the literature, cluster analysis is used more commonly to group symptom variables than to group patients [29]. The symptom variable clusters obtained are interpreted as syndrome types. This is problematic because symptom variables being strongly correlated with each other (and hence grouped together) does not necessarily imply that the symptoms tend to co-occur. Mutual exclusion also implies strong correlation. In addition, each symptom is placed in only one cluster and hence is related to only one syndrome type. In TCM theory, on the other hand, a symptom can be caused by multiple syndromes.

Factor analysis is another unsupervised learning method that has been used to quantify syndrome types in terms of symptoms [30]. Factor analysis assumes that observed variables (symptoms) are linear combinations of latent variables (syndrome types). The coefficients are called factor loadings. Researchers typically report factor loadings for latent variables and interpret the latent variables based on observed variables with high factor loadings. Factor analysis uses continuous latent variables, and hence it does not give patient clusters as the LTA method does. In addition, the factors are usually assumed to be mutually independent, while TCM syndrome types are correlated.

## 10. Conclusions

How to properly classify a population of patients into TCM syndrome types is a problem of fundamental importance to TCM research and clinic practice. The problem has not been satisfactorily solved before. A novel data-driven method for solving the problem is presented in this paper. It is called the LTA method and consists of six steps:

(1) Data collection: Conduct a cross-sectional survey of the population and collect data about the symptoms and signs of interest to TCM.

(2) Pattern discovery: Analyze data using latent tree models to reveal probabilistic symptom co-occurrence/mutual-exclusion patterns hidden in the data.

(3) Pattern interpretation: Interpret the patterns to determine their TCM connotations.

(4) Syndrome identification: Based on the patterns and their interpretations, identify a list of syndrome types that are well supported by the data.

(5) Syndrome quantification: Partition the population into patient clusters by performing joint clustering, match the patient clusters with syndrome types, and quantify the syndrome types using population statistics of the patient clusters.

(6) Syndrome classification: Derive a classification rule for each syndrome type from the results of syndrome quantification.

Among the six steps, steps 2, 5 and 6 are carried out using computers, while steps 1, 3 and 4 are done by TCM researchers.

Preliminary ideas behind the LTA method have appeared in the literature [31-38] in the past decade. Significant advances are made in this paper regarding the general framework, pattern interpretation, syndrome identification, syndrome quantification and syndrome classification. A software package called Lantern is developed to facilitate the use of the method. A complete case study is reported in an accompanying paper [1]. Researchers should be able to use the method in their own work after reading the two papers.

**Competing Interests**

The authors declare that they have no competing interests, financial or non-financial.

**Authors' contributions**

NLZ and CF were the main forces behind the conception and design of the work, and contributed equally to the paper. BXC and YLZ played key roles in data collection and results interpretation, and provided valuable comments on earlier versions of the manuscript. TFL, KMP and PXC developed the Lantern software that is released along with the publication of the paper, carried out the data analysis, and provided valuable comments on earlier versions of the paper. All authors have read and approved the final manuscript, and agree to be accountable to all aspects of the work.

**Acknowledgements:**

We thank Jerry Wing Fai Yeung for valuable comments on earlier versions of this paper. Research on this article was supported by Hong Kong Research Grants Council under grant 16202515, Guangzhou HKUST Fok Ying Tung Research Institute, China Ministry of Science and Technology TCM Special Research Projects Program under grants No.200807011, No.201007002 and No.201407001-8, Beijing Science and Technology Program under grant No.Z111107056811040, Beijing New Medical Discipline Development Program under grant No.XK100270569, and Beijing University of Chinese Medicine under grant No. 2011-CXTD-23.

syndrome types among patients with vascular mild cognitive impairment using latent tree analysis. Co-Submission.

2. Xu ZX, Zhang NL, Wang YQ, Liu GP, Xu J, Liu TF, and Liu AH. Statistical Validation of Traditional Chinese Medicine Syndrome Postulates in the Context of Patients with Cardiovascular Disease. *The Journal of Alternative and Complementary Medicine*, 2013, 18, 1-6.

3. Pearl J. *Probabilistic Reasoning in Intelligent Systems: Networks of Plausible Inference.* Morgan Kaufmann Publishers, San Mateo, California, 1988.

4. Zhang NL. Hierarchical latent class models for cluster analysis. *The Journal of Machine Learning Research*, 2004, 5:697–723.

5. Aldrich J. RA Fisher and the making of maximum likelihood 1912–1922. *Statistical Science*, 1991, 12 (3): 162–176.

6. Schwarz G. Estimating the dimension of model. *Ann. Statist.*, 1978; 6:461-464.

7. Bartholomew DJ, Knott M. *Latent variable models and factor analysis*, 2nd edition. Arnold, London, 1999.

8. Mourad R., Sinoquet C, Zhang NL, Liu TF, Leray P. A survey on latent tree models and applications. *Journal of Artificial Intelligence Research*, 2013, 47:157-203 .

9. Liu TF, Zhang NL, Chen PX, Liu AH, Poon LKM, Wang Y. Greedy learning of latent tree models for multidimensional clustering. *Machine Learning*, 2013:1-30.

10. Chen T, Zhang NL, Liu TF, Wang Y, Poon LKM. Model-based multidimensional clustering of categorical data. *Artificial Intelligence*, 2011, 176:2246-2269.

11. Zhang NL, Poon LKM, Liu TF, et al. Lantern software. http://www.cse.ust.hk/faculty/lzhang/tcm/. Accessed 18 January 2016.

**Table 1**. Partition of patients given by the latent variable Y06: The population is partitioned into two mutually exclusive clusters, consisting of 79% and 21% of the patients respectively. The partition is characterization using the symptom variables directly connected to Y06. Their occurrence probabilities in the two clusters are shown in columns two and three. The fourth column shows the mutual information between Y06 and the symptom variables.

|  | Y06=s0 (0.79) | Y06=s1 (0.21) | MI |
|---|---|---|---|
| Thick tongue fur | 0.05 | 0.63 | 0.16 |
| Greasy tongue fur | 0.38 | 0.79 | 0.06 |

**Table 2**. Partition of patients given by the latent variable Y20

|  | Y20=s0 (0.91) | Y20=s1 (0.09) | MI |
|---|---|---|---|
| Fat tongue | 0.05 | 0.58 | 0.08 |
| Tongue with ecchymosis | 0.02 | 0.30 | 0.04 |
| Tooth-marked tongue | 0.08 | 0.46 | 0.04 |

**Table 3**. Partition of patients given by the latent variable Y12

|  | Y12=s0 (0.43) | Y12=s1 (0.57) | MI |
|---|---|---|---|
| Slippery pulse | 0.85 | 0.16 | 0.26 |
| Thin pulse | 0.00 | 0.57 | 0.24 |

**Table 4**. Partition of patients given by the latent variable Y25

|  | Y25=s0 (0.64) | Y25=s1 (0.36) | MI |
|---|---|---|---|
| Insomnia | 0.16 | 0.78 | 0.20 |
| Dreamfulness | 0.23 | 0.83 | 0.18 |
| Flushed face | 0.10 | 0.03 | 0.01 |

**Table 5**. Joint cluster results for Phlegm-Dampness: The patients are divided into three clusters Z=s0, Z=s1 and Z=s2. The last two clusters are merged into one cluster Z=s12. The sixth columns shows mutual information (MI) and the seventh column shows cumulative information coverage (CIC).

|  | Z=s0 (0.42) | Z=s1 (0.44) | Z=s2 (0.14) | Z=s12 (0.58) | MI | CIC |
|---|---|---|---|---|---|---|
| Greasy tongue fur | 0.03 | 0.86 | 0.60 | 0.80 | 0.36 | 56% |
| Sticky or greasy feel in mouth | 0.05 | 0.18 | 0.62 | 0.29 | 0.09 | 70% |
| Slippery pulse | 0.27 | 0.67 | 0.39 | 0.60 | 0.07 | 75% |
| Urinary incontinence | 0.17 | 0.13 | 0.65 | 0.26 | 0.07 | 84% |
| Dizzy headache | 0.02 | 0.00 | 0.25 | 0.06 | 0.06 | 88% |
| Expectoration | 0.26 | 0.20 | 0.63 | 0.30 | 0.05 | 92% |
| Dizziness | 0.45 | 0.42 | 0.80 | 0.51 | 0.03 | 95% |

**Table 6.** Classification rule for Phlegm-Dampness (Z=s12). The shaded part can be deleted if one wants to simplify the rule.

|  | Score | Threshold | Accuracy |
|---|---|---|---|
| Greasy tongue fur | 7.1 |  |  |
| Slippery pulse | 2.1 |  |  |
| Sticky or greasy feel in mouth | 2.8 |  |  |
| Thick tongue fur | 1.5 |  |  |
| Dizzy headache | 1.8 |  |  |
| tooth-marked tongue | 1.0 |  |  |
| Fat tongue | 1.0 |  |  |
| Urinary incontinence | 0.6 | 3.7 | 0.967 |
| Dizziness | 0.4 | 3.9 | 0.965 |
| Thirst desire no drinks | 0.6 | 4.0 | 0.966 |
| Head feels as if swathed | 0.4 | 4.1 | 0.966 |
| Expectoration | 0.3 | 4.2 | 0.969 |
| Nausea or vomiting | 0.4 | 4.2 | 0.969 |
| Distending headache | 0.2 | 4.2 | 0.969 |
| Insomnia | 0.02 | 4.2 | 0.969 |
| Dreamfulness | 0.02 | 4.2 | 0.969 |

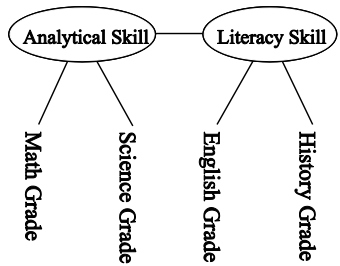

| P(MG\|AS) | MG=low | MG=high |
|---|---|---|
| AS=low | 0.8 | 0.2 |
| AS=high | 0.2 | 0.8 |

(a) A latent tree model

| P(AS) | AS=low | AS=high |
|---|---|---|
|  | 0.7 | 0.3 |

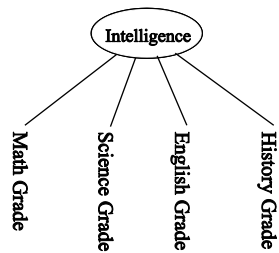

| P(LS\|AS) | LS=low | LS=high |
|---|---|---|
| AS=low | 0.6 | 0.4 |
| AS=high | 0.4 | 0.6 |

(b) A latent class model

AS = Aalytical Skill, MG = Math Grade, LS = Literacy Skill

**Figure 1.** The subfigure (a) and the tables illustrate the concept of latent tree models using an example that involves two latent variables (the skill variables) and four observed variables (the grade variables). The tables show some of the probability parameters for the latent tree model. The subfigure (b) illustrates the concept of latent class models where Intelligence is the only latent variable.

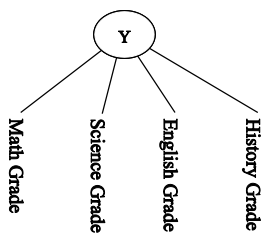

(a) The model for latent class analysis

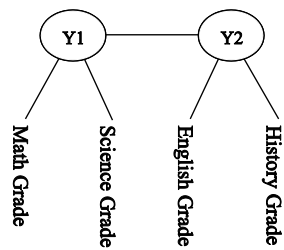

(b) A possible model from latent tree analysis

**Figure 2.** The subfigure (a) shows the model for latent class analysis. There is only one latent variable Y. The task is to determine the number of values for Y and the probability parameters. The subfigure (b) shows a model that might be obtained from latent tree analysis, where it is necessary to determine the number of latent variables and connections among them additionally.

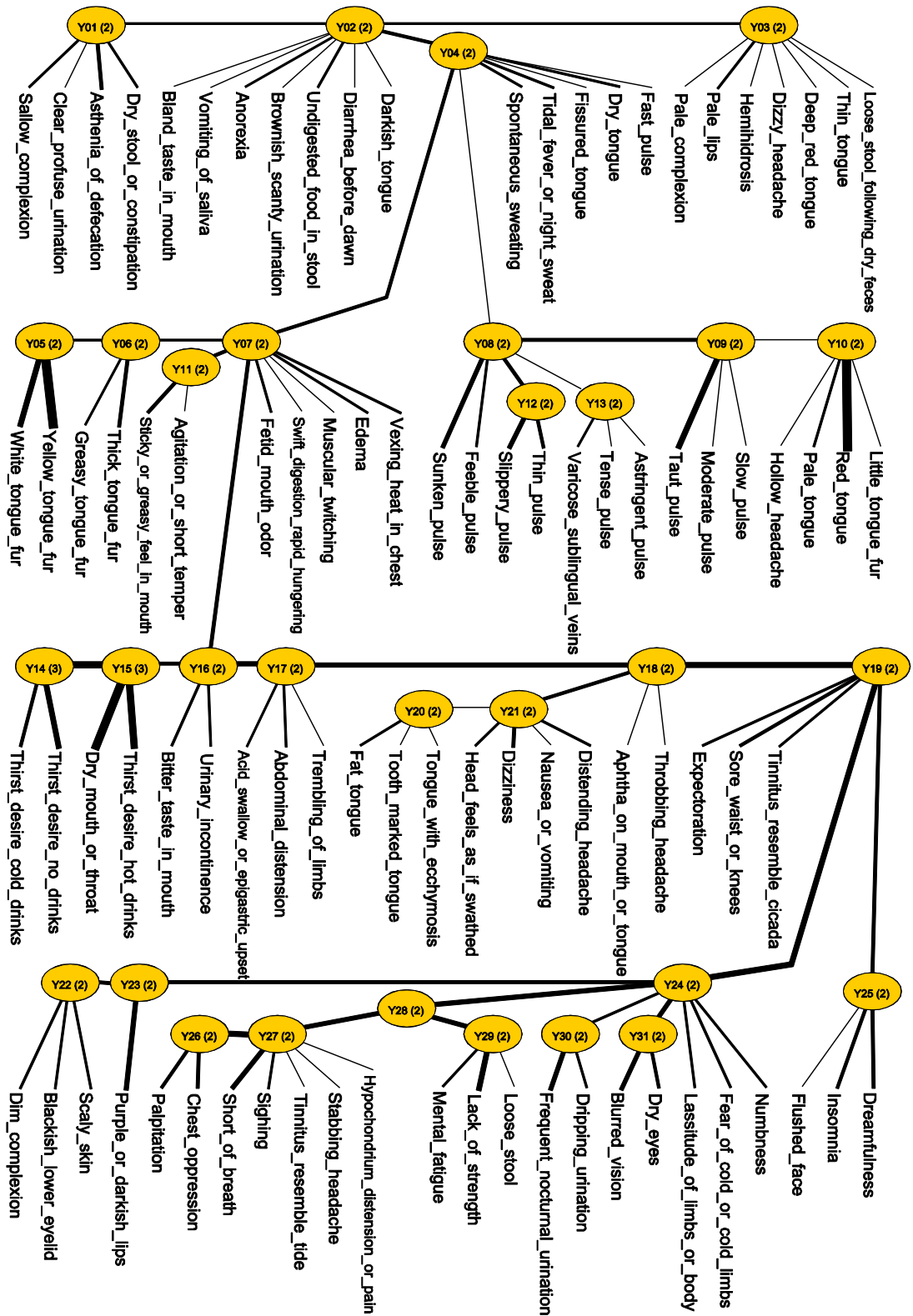

**Figure 3.** Structure of the latent tree model obtained on the VMCI data: The variables labeled with English phrases are symptom variables, while the Y-variables are latent variables. The integer next to a latent variable is the number of its possible values.

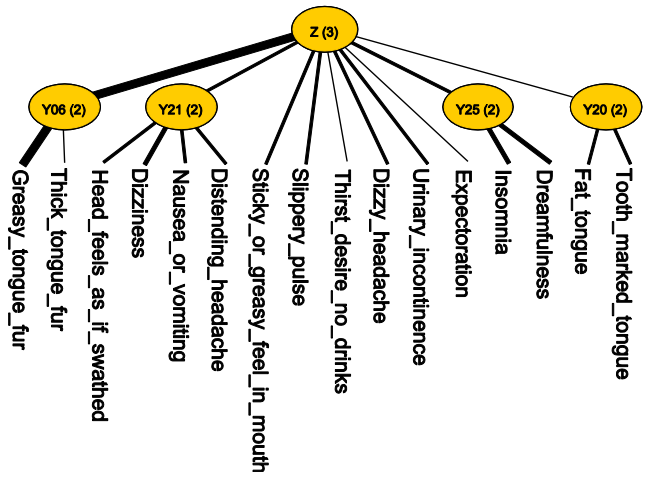

**Figure 4.** Joint clustering model for Phlegm-Dampness.